\RequirePackage{fix-cm}
\documentclass[twocolumn]{svjour3}          
\usepackage{graphicx}
\usepackage{xcolor}
\definecolor{wong-black}        {HTML}{000000}
\definecolor{wong-lightorange}  {HTML}{E69F00}
\definecolor{wong-lightblue}    {HTML}{56B4E9}
\definecolor{wong-green}        {HTML}{009E73}
\definecolor{wong-yellow}       {HTML}{F0E442}
\definecolor{wong-darkblue}     {HTML}{0072B2}
\definecolor{wong-darkorange}   {HTML}{D55E00}
\definecolor{wong-pink}         {HTML}{CC79A7}
\definecolor{gray-table}        {HTML}{000000}  
\definecolor{gray-background}        {HTML}{EAEAEA}

\PassOptionsToPackage{hyphens}{url}

\usepackage{array}
\usepackage{colortbl}
\usepackage{tablefootnote}
\usepackage{threeparttable}

\usepackage{cite}
\usepackage{siunitx}
\usepackage{booktabs}
\usepackage{bigstrut}
\usepackage{amsmath,amssymb,amsfonts}
\usepackage{algorithmic}
\usepackage{multirow,graphicx}
\usepackage{textcomp}

\usepackage[english]{babel}

\usepackage[strings]{underscore}
\usepackage{csquotes}
\usepackage{parnotes}
\usepackage{tabularx}
\usepackage{ifthen}
\usepackage[shortlabels]{enumitem}
\usepackage{hyperref}

\journalname{Preprint}

\begin{document}
	\sloppy 
	
	\title{Checklist to Define the Identification of TP, FP, and FN Object Detections in Automated Driving
		

	}
	%
	\titlerunning{Checklist to Define the Identification of TP, FP, and FN Object Detections in Automated Driving}        
	
	\author{Michael Hoss (corresponding author, ORCID 0000-0001-9924-7596)}
	
	\authorrunning{Michael Hoss} 

	\institute{RWTH Aachen University, Templergraben 55, 52062 Aachen, Germany.\\
		E-mail: michael.hoss@rwth-aachen.de}
	
	\date{Received: date / Accepted: date}
	
	\maketitle
	\begin{abstract}
				
		
		The object perception of automated driving systems must pass quality and robustness tests before a safe deployment.
		Such tests typically identify true-positive (TP), false-positive (FP), and false-negative (FN) detections and aggregate them to metrics. 
		Since the literature seems to be lacking a comprehensive way to define the identification of TPs/FPs/FNs, this paper provides a checklist of relevant functional aspects and implementation details. 
		Besides labeling policies of the test set, we cover areas of vision, occlusion handling, safety-relevant areas, matching criteria, temporal and probabilistic issues, and further aspects.
		Even though the checklist cannot be fully formalized, it can help practitioners minimize the ambiguity of their tests, which, in turn, makes statements on object perception more reliable and comparable.

		\keywords{Test Oracle \and Object Perception \and False Positives \and False Negatives \and Automated Driving \and Safety Assurance}
	\end{abstract}

	\section{Introduction}
	\label{sec:introduction}
	
	
	
	Test metrics such as HOTA \cite{Luiten2020hota} and the CLEAR MOT metrics \cite{bernardin2008evaluating} are commonly used to benchmark the object perception of automated driving systems (ADS).
	Such metrics typically consist of aggregations of true-positive (TP), false-positive (FP), and false-negative (FN) object detections. Based on data of TPs/FPs/FNs, test criteria can determine whether a test passes or fails. 
	For example, a trivial test oracle could equate failure with the presence of FPs/FNs.
	As this paper focuses only on the identification of TPs/FPs/FNs, it assumes failure upon FPs/FNs and leaves more complex criteria open to the reader.
	With the assumption of these simple pass/fail criteria, the task of the test oracle boils down to the identification of TPs/FPs/FNs in the context of this paper. 
	
	Test oracles have many degrees of freedom, including, but not limited to, areas of vision of the reference data, how the reference data are labeled, criteria for object matching, cutoffs beyond certain distances, exclusion of certain types of objects, temporal and spatial alignment of the data, and thresholds for confidence values.

	Since in practice, it is not straightforward to characterize these many degrees of freedom of a test oracle in full detail, test activities typically only state what is relevant in their given context. 
	Such oracle-defining statements can span across labeling specifications, the test equipment's user manuals, online documentations, sensor specifications, and the source code that implements the oracle. 
	Some aspects might not have explicit documentation at all, for example human processes for label quality assurance. 
	As long as the goal of testing is to provide feedback to developers on early-stage research, test oracles don't need to have all degrees of freedom transparently documented. 
	
	
	
	This changes when the goal of testing is to approve the release of ADS to public roads.
	When companies have their public testing permits suspended or revoked due to safety concerns \cite{Nacto2023_cruise_revocation, Roy2024_California}, 
	making test procedures transparent can reestablish trust in the technology.
	The so-called SMIRK safety case \cite{Borg2022smirk, Socha2022_SMIRK} already uses performance requirements based on TPs/FPs/FNs to argue the safety of a machine learning-based emergency braking system. 
	Usually, TPs are considered safe and FPs/FNs unsafe.
	However, in general, TPs can also be unsafe if their velocities are wrongly estimated and FPs/FNs can be safe if they are sufficiently far away. 
	Only with clear definitions of what a TP is in the first place, or what it means to be sufficiently far away, metrics based on TPs/FPs/FNs have a chance to form reliable evidences for claims about quality, robustness, and safety. 
	To provide more clarity in this regard, the present paper contributes:

	\begin{itemize}
		\item A detailed checklist of aspects that impact the identification of TPs/FPs/FNs (Sec. \ref{sec:criteria}) 
		\item Concrete examples of two test oracles for TPs/FPs/FNs expressed in terms of the checklist (Sec. \ref{sec:case_studies})
	\end{itemize}


	


		
		
		
	
	\begin{figure*}[t]
		\centering
		\vspace*{2mm}
		\includegraphics[width=\textwidth]{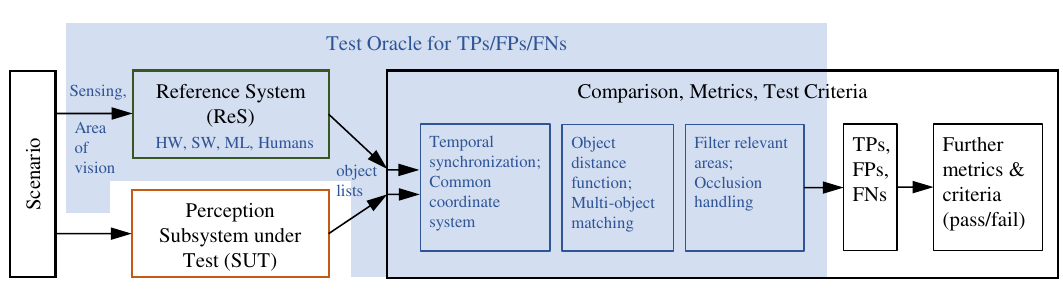}
		
		\caption{Identification of TPs/FPs/FNs, given a scenario and an SUT; using the taxonomy for ADS perception testing \cite{Hoss2022review, stellet2015testing}. In general, oracles may implement these modules in various orders.
		}
		\label{fig:oracle_in_taxonomy}
	\end{figure*}

	\subsection{Definition of terms \& acronyms}
	Unless stated otherwise, the following term definitions and acronyms hold throughout this paper.
	
	\emph{Object:} \label{def:object}
	From ISO 23150:2021 \cite{ISO_23150_2021_data_communication}: 
	``representation of a real-world entity with defined boundaries and characteristics in the vehicle coordinate system".
	This paper narrows down this definition to Layer 4 of the 6-layer-model \cite{Scholtes20216lmAccess}. An object can span over one or multiple time steps. 
	
	\emph{System under test (SUT)}: \label{def:sut} 
	Perception subsystem of an ADS. It comprises both hardware and software and it outputs an object list in each time step. 
	
	\emph{Reference system (ReS):}
	\label{def:reference_system}
	Perception system that observes the scenario and whose outputted object list serves as a desired reference for the SUT's object list. 
	
	\emph{Object matching:} \label{def:association} 
	As part of testing, finding corresponding objects between the SUT object list and the ReS object list.
	Not to be confused with object \textit{association}, which is a matter of a perception algorithm (finding correspondences between novel detections and existing tracks; adopted from \cite{Luiten2020hota}).
	
	\emph{True positive (TP):} \label{def:tp} The circumstance that an SUT object matches with a ReS object (Fig. \ref{fig:top_down_all}, A).
	
	\emph{False positive (FP):} \label{def:fp} The circumstance that an SUT object does not match with any ReS object (Fig. \ref{fig:top_down_all}, B). 
	
	\emph{False negative (FN):} \label{def:fn} The circumstance that no SUT object matches with a given ReS object (Fig. \ref{fig:top_down_all}, C).
	
	\emph{True negative (TN):} \label{def:tn} The circumstance that the absence of an SUT object matches with the absence of a ReS object. Since world modeling in form of object lists represents the absence of objects only implicitly, TNs are not quantifiable in the context of this paper.
	
	
	


	\emph{(Test) oracle:} \label{def:oracle} 
	In general: the mechanism that determines whether a test passes or fails. In the present paper: the mechanism that identifies TPs/FPs/FNs, where the traffic scenario and the SUT are given (Fig.~\ref{fig:oracle_in_taxonomy}).

	
	\section{Related Work}
	\label{sec:related_work}
	
	To provide a broader context, we review the role of TPs/FPs/FNs in current perception metrics (Sec. \ref{sec:related_work_context}). Based on this, we specifically review the definition of test oracles for TPs/FPs/FNs (Sec. \ref{sec:related_work_oracles}).


	\subsection{TPs/FPs/FNs in current perception metrics}
	\label{sec:related_work_context}
	Safety assurance of the overall ADS becomes simpler when metrics can detect perception events that induce vehicle-level failures \cite{Oboril2022mtbf_ieee}. 
	Traditional perception metrics such as HOTA or the CLEAR MOT metrics cannot identify these fail events because they don't consider an object's relevance for the driving task \cite{Madala2023metrics, Willers2020safety, Lyssenko2021relevance, Piazzoni2020Modeling_conference, Aravantinos2020}. 
	Also the ability of the mentioned metrics to rank algorithms in a leaderboard has been shown to fluctuate unreasonably with parameter choices \cite{Nguyen2022_How}. 
	Consequently, novel perception metrics are being proposed, often with the intention to be task-oriented, planner-centric, and/or safety-relevant. Most of them explicitly identify TPs/FPs/FNs \cite{Volk2020metric, Lyssenko2021relevance,  Lyssenko2022safety, Wolf2021people, Bansal2021riskrankedrecall, Topan2022zones, 
		Ivanovic2021_Injecting, Ceccarelli2023_Evaluating}, while other approaches are independent of TPs/FPs/FNs \cite{Philion2020planner_centric, Li2023_Transcendental, shalevshwartz2017formalRSS}.
	In this context, Schreier et al. \cite{Schreier2023_Offline} demonstrated how planner-centric metrics should not be relied on exclusively as long as planner-independent metrics based on TPs/FPs/FNs still predict driving performance best.
	
	
	In summary, current perception metrics are expected to not only give helpful feedback to developers, but also indicate whether safe vehicle operation is possible. While this responsibility makes metrics more complex and thereby potentially harder to comprehend, the proposed clearer identification of TPs/FPs/FNs brings more clarity back into the metrics.
	In particular, isolated perception metrics better comparable and derived task-oriented metrics become more dependable. 
	

	\subsection{Test oracles for TPs/FPs/FNs}
\label{sec:related_work_oracles}


The explicit definition of test oracles for ADS appears to be only sparsely represented in the public literature. 
We are aware of Abrecht et al. \cite{Abrecht2021testing}, who discuss test oracle generation for visual perception by focusing on functional properties of deep neural networks (DNNs). 
The survey by Tang et al. \cite{Tang2023_Survey} reviews different approaches for test oracle generation of ADS perception, which are ground truth labeling, metamorphic testing, and formal specifications. 
Further literature on test oracles is targeted towards the driving quality of ADS \cite{Jahangirova2021_Quality} or software in general \cite{Barr2015oracle}.
None of the mentioned works cover practical implementation aspects of oracles for TP/FP/FN object detections. 


	\section{Methods} 
\label{sec:method}

In the present context, an oracle obtains a reference object list and compares it to the SUT's object list (Fig.~\ref{fig:oracle_in_taxonomy}). 
We compiled the following checklist (Sec.~\ref{sec:criteria}) to help define, but not prescribe, oracles more transparently. 
The following aspects and considerations influenced the checklist:
\begin{itemize}
	\item human comprehensibility of the checklist's aspects \newline (to keep testing procedures understandable)
	\item usefulness in different stages of development \newline (e.g. initial developer feedback and final evidences for quality and robustness)
	\item application to different focused parts of the SUT (e.g. DNNs or sensor hardware)
	\item the literature on task-oriented perception testing \newline (to incorporate the academic state of the art)
	\item being able to define actual oracles (for usability and to incorporate the practical state of the art)
	\item own experiences of what matters in an oracle

\end{itemize}



\section{Checklist to Define the Identification of TPs/FPs/FNs}
\label{sec:criteria}

\begin{figure*}[t]
	\centering
	\vspace*{2mm}
	\includegraphics[width=\textwidth]{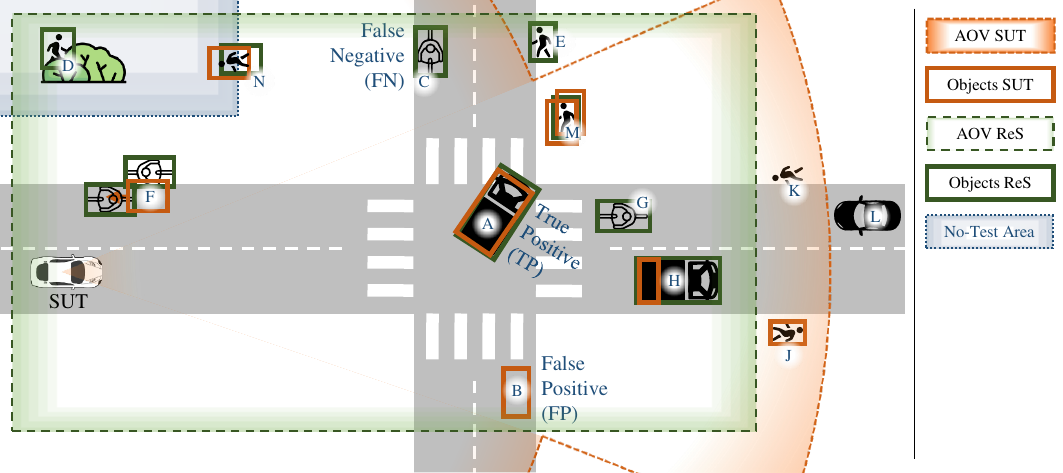}
	
	\caption{ While objects A-C appear obvious, the desired identification of TPs/FPs/FNs from objects D-N requires a purposefully defined test oracle (best viewed in color). 
	}
	\label{fig:top_down_all}
\end{figure*}

A test oracle can be regarded from different perspectives: as part of a testing activity (Fig. \ref{fig:oracle_in_taxonomy}), in space (Fig. \ref{fig:top_down_all}), and in time (Fig. \ref{fig:timeline}).
Each paragraph of this section first highlights certain degrees of freedom of the oracle and then provides a checklist to transparently define these degrees of freedom.

After the simple case of deterministic object representations in a single time frame (Sec. \ref{sec:oracle_simple}), temporal aspects (Sec. \ref{sec:oracle_time}) and probabilistic data representations (Sec. \ref{sec:oracle_probabilistic}) are covered.



\subsection{Deterministic objects in a single time frame}
\label{sec:oracle_simple}

In this subsection, we assume that the object lists of SUT and ReS are perfectly temporally synchronized. 

\subsubsection{Area of vision (AOV) of ReS}
\label{sec:basic_aov_ref}
Different AOVs of SUT and ReS can lead to undesired FNs and FPs in areas without overlap.
In Fig. \ref{fig:top_down_all}, the SUT has no chance to detect pedestrian E because of insufficient sensor range at this radial position.
Likewise, the ReS has no chance to detect pedestrians K and J, or car L, given its AOV. 

Differing AOVs can have multiple reasons such as different perspectives (e.g. a camera-equipped drone as ReS hardware in  Fig.~\ref{fig:top_down_all}), different choices of sensor modalities, or different ranges up to which the labeling process reaches.


Whether or not the SUT should be penalized for missing pedestrian E in Fig. \ref{fig:top_down_all} can depend on the test objective. 
If only the perception software is tested, it would be unfair to identify a FN, but otherwise, a FN would reveal the hardware-induced AOV limits.


\paragraph{Aspects to define}
\begin{itemize}
	\item Geometry of the ReS AOV generally and in each time step
	\item Geometric relation of ReS AOV to SUT AOV
\end{itemize}

\subsubsection{Perspective-related occlusions}
\label{sec:basic_occlusions}

If SUT and ReS observe the scenario from different perspectives, certain traffic participants can be visible for one system, but occluded for the other, which can lead to FPs and FNs. 
In Fig.~\ref{fig:top_down_all}, truck A occludes both cyclist G and pedestrian K for the SUT. 
Cyclist G, which is still within the ReS AOV, may be identified as a FN or may be excluded from evaluation, depending on whether the SUT is expected to see behind such occlusions.
This would be possible by temporarily maintaining its track or by radar beams below truck~A. 
There might as well be no such expectations if the downstream behavior planning module assumes a road user behind occlusions anyway.

\paragraph{Aspects to define}
\begin{itemize}
	\item Areas occluded to the ReS
	\item Areas visible to the ReS, but occluded to the SUT, and if the SUT is being tested there 
	\item What counts as an occlusion (depending on degree of occlusion, who or what blocks the vision etc.)
\end{itemize}

\subsubsection{ReS hardware}
\label{sec:basic_ref_hw}

In the simplest case, ReS and SUT share the very same sensor hardware and just differ in their data processing. 
This case would exclude unexpected FPs and FNs based on hardware differences, which is desired if only the SUT's software/data processing is focused. 

However, also the SUT's hardware can be put under test by utilizing distinct ReS sensors for identifying the SUT's hardware limitations in, for example, adverse weather conditions. 

\paragraph{Aspects to define}
\begin{itemize}
	\item Absolute sensor properties of the ReS (resolution, performance in relevant adverse environmental conditions etc.)
	\item Whether/how the ReS sensor hardware is superior to the SUT sensor hardware 
\end{itemize}

\subsubsection{ReS labeling and data processing}
\label{sec:basic_ref_labeling}

The method by which an object list is extracted from the raw reference data 
can comprise software, machine-learned models, and human interaction. 
Labeling policies are commonly specified to create consistency among ReS objects, which furthermore helps to clarify expectations in perception challenges and for outsourcing data labeling to subcontractors. 
These labeling policies include details such as how objects should be classified, criteria for including and omitting objects, which parts of objects shall be inside or outside their bounding boxes, or how to label in case of partial or full occlusions. 

To align with requirements from downstream, the mentioned criteria can be specified such that FPs/FNs represent potential causes of vehicle-level failures. 
However, encoding a full prior specification of necessary conditions for safe driving into the TP/FP/FN oracle is likely impossible in the highly complex open context of public roads \cite{poddey2019opencontext, Salay2019partialspecifications}. 

Even with a thorough labeling policy, the actual labels can be uncertain \cite{wang2020inferring}, erroneous, or part of underspecified edge cases.
To still describe the ReS labeling as transparently as possible, one could additionally describe how the labeling policy is implemented in practice.
This includes
the programmed modifications that the data undergo, the exact machine learning processes of any machine-learned modules, and the human interventions (when and how do humans manually label or otherwise interact with the data).
More clarity can also be provided by statistics of labeling accuracy and precision using a higher-level reference system or evaluation procedure (``reference of the reference").

\paragraph{Aspects to define}
\begin{itemize}
	\item Criteria as exact as feasible for labeling ReS object state, classification, and existence
	\item Qualitative issues impacting label quality related to human, machine-learned and programmed data manipulations  
	\item Statistics about the imperfections of the ReS labels
\end{itemize}

\subsubsection{Relevant areas and no-test areas}
\label{sec:basic_areas}

While certain areas on or next to the road surface are relevant to the traffic scenario, others might not be worth while perceiving. 
To allocate test efforts meaningfully, one can therefore define (safety-) relevant areas and no-test-areas \cite{Philipp2022systematization, Topan2022zones, Butz2020soca, Wolf2021people, Chu2023sotif}.
Similarly, \cite{Mori2023relevance} and \cite{Storms2023relevance} determine directly which objects are relevant for perception.

The exclusion of objects from no-test areas (e.g. pedestrian D in Fig.~\ref{fig:top_down_all}) can occur in different stages of the testing activity (Fig. \ref{fig:oracle_in_taxonomy}). 
They might already be excluded from the reference data; they might be included in the reference data, but excluded from the identification of TPs/FPs/FNs; or, they could still be part of the TPs/FPs/FNs, but excluded in further metrics and criteria.

In any case, a clear definition of such areas is crucial for a comprehensible test oracle (e.g. farther away than \SI{20}{\meter} from any road surface; or above/below any other threshold).
The definition of no-test areas can be delicate due to the risk of omitting objects that are in fact relevant. 
For example, the excluded pedestrian D in Fig.~\ref{fig:top_down_all} might still be running towards the intersection and thereby influence other trajectories.

\paragraph{Aspects to define}
\begin{itemize}
	\item General criteria to include/exclude certain areas or objects from testing
	\item Concrete boundaries between included and excluded areas/objects in each time step
\end{itemize}

\subsubsection{Coordinate transformation from ReS to SUT}
\label{sec:basic_spatial_alignment}

While certain oracle aspects such as relevant areas can be seen as functional properties, the present aspect is rather a non-functional matter of implementation.
If ReS and SUT use different coordinate systems, then the ReS objects' coordinates must be transformed to the SUT's coordinate system before matching can take place.
Since uncertainties in this transformation can ultimately affect the matching of object pairs, knowledge of the uncertainty inherent in this geometrical alignment is crucial. 


Especially, uncertainties in the sensor orientation angles result in large cartesian offsets far away from the sensor.
For reference sensors on the ego vehicle, such uncertainties could stem from the calibration of sensor mounting positions. 
For external reference sensors such as moving drones or mounted infrastructure sensors, the geometrical alignment has further challenges. 
This includes a generally uncertain localization of the ego vehicle and of the external sensor (using RTK-GNSS, an offline map, and/or perceiving each other), and a transformation from the aerial perspective to the SUT's coordinate system over the ground. 

\paragraph{Aspects to define}
\begin{itemize}
	\item Error of ReS object positions resulting from imperfect transformation to SUT coordinates
	\item Aspects and mechanisms that cause this error
\end{itemize}

\subsubsection{Object distance function for matching}
\label{sec:basic_dist_func}

Once both object lists share the same coordinate system, pairwise distances between ReS objects and SUT objects can be computed to find candidates for TPs. 
Commonly used distance functions are the euclidean distance between bounding box centers in the nuScenes tracking task \cite{nuScenesTracking2024} 
or $1 - IoU$ in KITTI \cite{Kitti2024_bev_eval}
in either 2D bird's eye view (BEV) or 3D. 


While the bounding boxes of truck A (Fig.~\ref{fig:top_down_all}) should match for any reasonable choice of distance function and threshold value, the situation is less obvious for truck H.
Since only its rear end got detected by the SUT, both the center distance and $1 - IoU$ could prevent a match even with common thresholds like \SI{2}{\metre} or $0.5~IoU$, respectively. 
Without a match, though, truck H would lead to a FP and a FN, which might be undesired given the fact that its property that is relevant to the driving task of the ego vehicle, namely its rear end, got properly detected. 
Still, thresholds should also not be unnecessarily loose to prevent unrelated misdetections such as G and B from merging into an undesired TP candidate.


Furthermore, an object pair that falls below the distance function's threshold does not automatically represent a TP. 
For example, the SUT bounding box of cyclist F (Fig.~\ref{fig:top_down_all}) could have a below-threshold centroid distance to both ReS bounding boxes nearby. 
To disambiguate the TP object pair, penalty terms such as the mismatch in estimated velocity or yaw/heading angle could be added to the distance function.
An oracle could also model differing road user classes as a (match-preventing) penalty term within the distance function for matching. 


\paragraph{Aspects to define}
\begin{itemize}
	\item Geometrical distance function and threshold for matching ReS objects to SUT objects
	\item Penalty terms or criteria for preventing matches 
\end{itemize}

\subsubsection{Multi-object matching algorithm}
\label{sec:basic_multi_obj_matching}

With all pairwise object distances given, a multi-object matching algorithm can finally identify the resulting TPs/FPs/FNs. 
Often, algorithms only match one SUT object with one ReS object (1:1), but 1:n, n:1, or n:n matches could also be allowed \cite[Sec. 11.3]{Brahmi2020diss}. 
For cyclists F in Fig.~\ref{fig:top_down_all}, the algorithm would match the SUT bounding box to the ReS bounding box with the closest distance function value as long as the distance is below the threshold (Sec. \ref{sec:basic_dist_func}). 
With n:n matching and both distances below the threshold, there would be no FN in F. 
Likewise, pedestrian M is detected twice by the SUT, which would result in one FP if only 1:1 is allowed. 

While the object list matching in Fig.~\ref{fig:top_down_all} is straightforward, there may also be complex scenes like a dense crowd of pedestrians, where each bounding box has many potential matches. 
In such cases, optimization algorithms such as the Hungarian/Munkres algorithm are needed to determine TPs by globally minimizing the overall distance between all matches.


\paragraph{Aspects to define}
\begin{itemize}
	\item Algorithm to disambiguate nontrivial matches
	\item Allowed cardinality of matches (1:1, 1:n, n:1, n:n)
\end{itemize}

\subsubsection{Corner cases of different degrees of freedom}
\label{sec:basic_corner_cases}

Individual degrees of freedom of a test oracle might contradict each other in ways that need explicit resolution. 
For example, for pedestrian N (Fig. \ref{fig:top_down_all}), the objects of ReS and SUT are close enough for a match (\ref{sec:basic_dist_func}), but at the border of the tested area (Sec. \ref{sec:basic_areas}), the ReS object is included while the SUT object is excluded. 
In this case, the oracle could resolve to either a hard cut resulting in a FN, or a fuzzy cut that still considers nearby objects in the no-test-area if a TP can be obtained in this way.

\paragraph{Aspects to define}
\begin{itemize}
	\item Unambiguous resolutions to potentially contradictory rules of the oracle
\end{itemize}

\subsection{Generalization to object representations over time}
\label{sec:oracle_time}

\begin{figure}[t]
	\centering
	\vspace*{2mm}
	\includegraphics[width=0.49\textwidth]{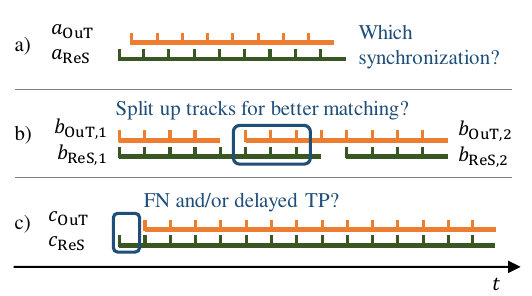}
	\caption{Temporal design aspects of a test oracle for TPs/FPs/FNs.
	}
	\label{fig:timeline}
\end{figure}

This section refers to Fig.~\ref{fig:timeline}, which contains pairs of tracks that would match geometrically according to the criteria from the previous section (Sec. \ref{sec:oracle_simple}). 
The vertical ticks denote the systems' sampled time frames.
Now, the test oracle's temporal aspects, which are explained in the following, determine how these objects match in time. 


\subsubsection{Synchronization of measurement time stamps}
\label{sec:temp_sync}

If the recorded ReS sampling times do not coincide with the SUT sampling times, synchronization of both systems must happen offline during post-processing. 
Continuously present reference tracks, e.g. from RTK-GNSS recordings, can be linearly interpolated onto the SUT time stamps \cite[Sec. 10.2.7]{Brahmi2020diss}.

However, in general, reference tracks from camera, lidar, or radar data do not span continuously over the entire recording, which means that the synchronization must also deal with track beginnings and ends. 
If the sampling times of track $a_\text{ReS}$ (Fig. \ref{fig:timeline}) are linearly interpolated to those of $a_\text{SUT}$, all sampling times of $a_\text{SUT}$ are covered. 
Then, however, it is up to the test oracle to decide what happens to the closest overhanging frames of $a_\text{ReS}$ at the beginning and at the end (see also Sec.~\ref{sec:temp_latency}). 
Both could be neglected, but they could also both mark FNs. 
Whether or not the test oracle identifies FNs there might also depend on whether the overhang falls below a temporal matching threshold. 


\paragraph{Aspects to define}
\begin{itemize}
	\item Sample times of ReS recording relative to SUT
	\item Algorithm for synchronizing/interpolating frames at the beginning, in the middle, and at the end of track lifetimes
\end{itemize}

\subsubsection{Matching of objects with an extended lifetime}
\label{sec:temp_matching}

Depending on the perception task whose output is being tested (e.g. detection, filtering, tracking), the tested SUT objects may or may not contain IDs to indicate their correspondence over consecutive time frames. 
For completeness, we assume here that the ReS objects do offer track IDs. 
The following representations of SUT objects can be matched to ReS objects:
\begin{itemize}
	\item \textit{detections or ID-stripped single frames out of tracks} (boils down to testing individual decoupled time frames, as explained in Sec. \ref{sec:oracle_simple}, and applied in e.g. the original OSPA metric \cite{schuhmacher2008consistent}) 
	\item \textit{tracks} (entire SUT and ReS tracks are matched with each other without being split up)
	\item \textit{sub-sequences of tracks} (tracks are split up into better-matching sub-sequences, as done in e.g. the CLEAR MOT metrics \cite{bernardin2008evaluating})
\end{itemize}

For example, if only individual frames of tracks are matched irrespective of their belonging to a track, then example b) (Fig.~\ref{fig:timeline}) would contain exactly one FN and one FP. 
If entire tracks are matched, then $b_{\text{SUT},1}$ would match with $b_{\text{ReS},1}$ and $b_{\text{SUT},2}$ would match with $b_{\text{ReS},2}$.
These matches would, however, have large temporal overhangs.
To reduce the extent of these overhangs, one could match also sub-sequences of tracks, such as the latest three frames of $b_{\text{ReS},1}$ with the first three frames of $b_{\text{SUT},2}$.
An ID switch could then be marked for $b_{\text{ReS},1}$ to still penalize the SUT's fragmented tracking.

\paragraph{Aspects to define}
\begin{itemize}
	\item Rules for the lifetime of a match (e.g. single frame or partial/entire tracks)
\end{itemize}

\subsubsection{Missed frames and incomplete tracks}
\label{sec:temp_incomplete}

Track $b_{\text{ReS},1}$ (Fig.~\ref{fig:timeline}) is perceived by the SUT in all but one frame, namely the frame in between $b_{\text{SUT},1}$ and $b_{\text{SUT},2}$. 
The test oracle could naturally identify a FN in this missing frame. 
However, if a single missed frame in the middle can be proved to be irrelevant to the driving task, a task-oriented test oracle might also intentionally not want to identify this missed frame as a FN. 


\paragraph{Aspects to define}
\begin{itemize}
	\item Penalization or allowance of missed individual frames of an object
\end{itemize}

\subsubsection{Treatment of latency and delays}
\label{sec:temp_latency}

Test methods may date back the object lists of SUT and ReS to the time stamps of their raw data acquisition. 
This approach is straightforward, but prevents the test results from reflecting the impact that computation time has on overall perception performance for real ADS. 

To account for the SUT's computation time, the test method would need to date back only the ReS objects, but stamp the SUT objects at the time when they become available to the subsequent behavior module. 
In this way, computation time of the SUT can be considered explicitly by TP properties such as the average delay metric \cite{Mao2019delay}, but also implicitly by initial FNs. 
Furthermore, the test oracle can distinguish FNs caused by long computation times from FNs caused by functional failures if it is aware of the SUT's computation time. 


Similar to the time for computation at the birth of a track, also the time to fade out a disappearing object at the end of a track can affect TPs/FPs/FNs. Oracles may have a policy for identifying or preventing FPs/FNs at the end of matched tracks if the SUT track ends slightly sooner or later than the ReS track.

\paragraph{Aspects to define}
\begin{itemize}
	\item Which stages of data processing provide the time stamps for synchronizing ReS and SUT objects
	\item Degree to which the SUT's computation time is reflected in the obtained TPs/FPs/FNs
	\item Temporal uncertainties (delays in ReS objects, uncertain time stamping)
\end{itemize}

\subsection{Generalization to probabilistic object representations}
\label{sec:oracle_probabilistic}

While the previous two sections assumed deterministic representations of objects and AOV, the following subsections highlight the test oracle's degrees of freedom for evaluating probabilistic data.

\subsubsection{Probabilistic AOVs}
\label{sec:prob_for}

Typically, the performance of active sensor systems degrades gradually with increasing distance or angle, rather than facing a sharp drop from 100\% to 0\% somewhere.
This behavior can be described by probabilistic maps for the capability of object detection in border areas of the AOV. 
With such maps, a clear distinction of intersecting and overhanging AOVs of SUT and ReS (Sec. \ref{sec:basic_aov_ref}) is not possible unless sharp borders are introduced artificially by the test oracle.
With fuzzy borders, FPs and FNs are possible in regions where one system might randomly still detect an object that the other system might not detect (even though both systems might work perfectly in closer proximity to the sensor).

\paragraph{Aspects to define}
\begin{itemize}
	\item Modeling of AOV regions of ReS and SUT where unreliable performance is expected
	\item Method of identifying or neglecting TPs/FPs/FNs in unreliable regions
\end{itemize}

\subsubsection{Probabilistic object state and distance functions}
\label{sec:prob_bbox}
Uncertainties in an object's state can be described by probability distributions (both for SUT and ReS objects). 
There might be cases where a ReS object and an SUT object are deterministically too far away from each other for a match, but the overlap of their probability distributions in state space might still lead to a reasonable match under a probabilistic distance function.

The dissimilarity of two probability distributions for the bounding box centroids can be expressed by the Wasserstein or earth mover's distance.
A probabilistic version of the \textit{IoU} for rotated 3D bounding boxes is the \textit{JIoU} \cite{Wang2020inferring_iros}.
If object distance shall be expressed in terms of standard deviations of a given uncertainty, the Mahalanobis distance can be used. 
In the mentioned distance functions, probabilistic functions to evaluate classification accuracy such as the negative log-likelihood (NLL) could serve as penalty terms.

\paragraph{Aspects to define}
\begin{itemize}
	\item Modeling of state uncertainties in object matching
	\item Impact of ReS object state uncertainty on object matching 
\end{itemize}

\subsubsection{Probabilistic object existence and classification}
\label{sec:prob_thresholding}


SUTs typically compute confidences for object existence and/or classification.
A simple test oracle can threshold the existence confidence at a fixed value and assume the class of maximum classification confidence.
If the confidences of all classes are considered relevant for a given object, the Brier Score or the NLL can be used as criteria to identify TPs, where thresholds may depend on scenario properties.

Detected, but misclassified objects could either be identified as a TP with a property ``assigned wrong class", but also as a FN for the ReS class and a FP for the SUT class.




\paragraph{Aspects to define}
\begin{itemize}
	\item Criterion for allowing object matches under uncertain confidences for existence and classification
\end{itemize}

\section{Application to Concrete Test Oracles}
\label{sec:case_studies}


The checklist criteria from the previous section (Sec.~\ref{sec:criteria}) are illustrated by two concrete test oracles in Table~\ref{table:case_study}, namely the evaluation of the nuScenes tracking task \cite{caesar2019nuscenes} and our own previous test activity \cite{Krajewski2020UsingDrones}.

Both oracles can be characterized fairly well by the same proposed criteria despite their different purposes (ranking of submissions in a leaderboard vs. perception error modeling) and despite their different reference systems (human labeling on ego vehicle sensor data vs. automated extraction from a hovering drone).

The ``\textit{Not published}" labels for our own previous test oracle illustrate the tendency to omit relevant aspects of the oracle if no conscious effort is made in this regard.

Despite having followed the checklist, the level of detail of Table~\ref{table:case_study} is still not sufficient for a fully transparent oracle definition and leaves some details unclear. 
For example, most details of ReS labeling are omitted due to space reasons. 
The referenced external documents that 
contain such details should be regarded as fully part of the definition.




\newcommand{\scenariosDrone}{\textcolor{gray-table}{Ego vehicle statically observes an urban intersection from a sidewalk.}}
\newcommand{\outDrone}{\textcolor{gray-table}{Single lidar sensor with integrated object detection and tracking}}
\newcommand{\furtherDrone}{\textcolor{gray-table}{Modeling of how SUT deviates from ReS regarding TP object states}}

\newcommand{\scenariosChallenge}{\textcolor{gray-table}{Ego vehicle dynamically participates in various urban traffic scenarios.}}
\newcommand{\furtherChallenge}{\textcolor{gray-table}{Computation of metrics such as AMOTA and AMOTP \cite{weng2019baseline} for the online leaderboard}}
\newcommand{\outChallenge}{\textcolor{gray-table}{Submitted detection and tracking algorithms that operate on either camera-only, lidar-only, or all raw sensor data}}

\newcommand{\scenariosCol}{\textcolor{gray-table}{Scenarios}}
\newcommand{\sutCol}{\textcolor{gray-table}{SUT}}
\newcommand{\furtherCol}{\textcolor{gray-table}{Further metrics \& criteria}}

\newcommand{\basicChallengeAOV}{See \cite["Data collection"]{nuScenes2024_mainpage}: ReS sensor mounting positions, ranges, and opening angles on ego vehicle are numerically defined or graphically illustrated. ReS AOV $\supseteq$ SUT AOV.}
\newcommand{\basicChallengeOcclusion}{ReS suffers from ego-perspective occlusions. For a camera-only SUT, the ReS lidar might see beyond the camera SUT; SUT is still tested there. Visibility is binned into 4 percentage-defined bins. }
\newcommand{\basicChallengeReSHW}{See \cite["Data collection"]{nuScenes2024_mainpage} for sensor properties. ReS HW (cameras, lidar, radar, GNSS+IMU) $\supseteq$ SUT HW. No adverse weather.}
\newcommand{\basicChallengeReSLabling}{See human annotator instructions \cite{nuScenes2024_annotator}: they are detailed, but leave uncertainties (when ``reasonably sure" about object shape and location, use ``best judgment" on bounding box properties). ReS objects require 1 lidar or radar point. ``Multiple validation steps" \cite{caesar2019nuscenes}, but no official information or statistics on label accuracy. 
}
\newcommand{\basicChallengeAreas}{No evaluation beyond time-constant class-specific maximum detection and tracking ranges (\SI{40}{\meter} to \SI{50}{\meter}); no evaluation of bikes and motorcycles inside bike racks; list of excluded object classes.}
\newcommand{\basicChallengeGeometrAlign}{No error estimation given. Source of error: calibration of lidar, camera, and radar w.r.t. ego coordinate system. ReS-SUT-alignment is optimal if same sensors are used.}
\newcommand{\basicChallengeObjDistance}{2D center distance on the ground plane; threshold of \SI{2}{\meter}; a match requires equal object classifications.}
\newcommand{\basicChallengeMultiObjMatching}{Metrics from \cite{Bernardin2006mot_metrics} use 1:1 matching and Munkres' algorithm (see \cite{bernardin2008evaluating}).}
\newcommand{\basicChallengeCornerCases}{None specified.}

\newcommand{\tempChallengeSync}{Individual sample time stamps available for each sensor modality. Online synchronization of lidar (\SI{20}{\hertz}) and camera (\SI{12}{\hertz}), but evaluation only on keyframes (\SI{2}{\hertz}) where the sensors coincide. Radar operates on \SI{13}{\hertz} \cite{nuScenes2024_mainpage}. ReS tracks and SUT tracks are linearly interpolated ``to avoid excessive track fragmentation from lidar/radar point filtering" \cite{nuScenesTracking2024}. Behavior at track beginnings/ends unspecified.}
\newcommand{\tempChallengeMatching}{Metrics from \cite{Bernardin2006mot_metrics} keep a match over subsequent time steps for as long as it is below the threshold, even if another, new match candidate would have less cost.}
\newcommand{\tempChallengeIncomplete}{All missed frames are counted as FNs; additionally, the LGD (average longest gap duration in seconds) is computed.}
\newcommand{\tempChallengeDelays}{Synchronization based on time stamps of raw data acquisition. SUT's computation time is irrelevant to oracle, but can be additionally self-reported as FPS (frames-per-second of tracker). Temporal overhangs are FPs/FNs; additionally, the TID (average track initialization duration in seconds) is computed. ReS time stamps are subject to synchronization uncertainty.}

\newcommand{\probChallengeAOV}{AOVs are not modeled probabilistically. Strict cutoff (see above).}
\newcommand{\probChallengeObjDistance}{Deterministic distance function only}
\newcommand{\probChallengeExistClass}{ReS existence, SUT existence, and ReS classification are modeled track-constant and deterministic, whereas SUT existence is expected $\in [0,1]$ in each time step for the respective class. Matching only within the same nuScenes tracking class.
40 different existence thresholds are used to compute averaged metrics \cite{weng2019baseline}. 
}

\newcommand{\basicDroneAOV}{ReS AOV is rectangular (about \SI{85}{\meter} $\times$ \SI{45}{\meter}), whereas SUT AOV is \SI{145}{\degree} wide and $\approx$\,\SI{80}{\meter} long. 
	The AOVs overlap in the analyzed area.}
\newcommand{\basicDroneOcclusion}{ReS has no relevant occlusions and can see beyond the SUT's ego-perspective occlusions. \textit{Not published:} SUT objects were only analyzed in those time steps when they were not occluded (2D line of sight to SUT object completely free).
}
\newcommand{\basicDroneReSHW}{Camera-equipped drone with 4K resolution hovering in a steady state disturbed by wind ($\SI{2}{\centi\meter}/\text{px}$ on ground level). SUT's range accuracy $<\SI{10}{\centi\meter}$. No adverse weather.}
\newcommand{\basicDroneReSLabling}{Automated offline post-processing as in the inD dataset \cite{Bock2020ind} (here, only for cars). 
	Human annotators manage the ReS training, not the release of ReS labels. Label accuracy impacted by geometrical alignment (Sec. \ref{sec:basic_spatial_alignment}), temporal synchronization (Sec. \ref{sec:temp_sync}), and bounding box accuracy (``pixel level").  No accuracy statistics.}
\newcommand{\basicDroneAreas}{Included: cars on roads up to non-specified distances from the intersection center. Excluded: any non-road area, any non-car object.}
\newcommand{\basicDroneGeometrAlign}{Accuracy loss of order $\approx$\,\SI{0.1}{\meter} is not further calculated.
Alignment by a fitted 3\textsuperscript{rd} order polynomial from ReS to SUT coordinates, which is fitted through corresponding pixels in an orthophoto (reference) and a stabilized drone image (contains ego vehicle with SUT)}
\newcommand{\basicDroneObjDistance}{2D center distance on the ground plane; threshold of \SI{1.5}{\meter}; only car-car matching.}
\newcommand{\basicDroneMultiObjMatching}{Entire tracks are matched (more details under temporal aspects). \textit{Not published:} All candidate SUT tracks are matched to a respective ReS track (also at the same time; SUT-ReS n:1).}
\newcommand{\basicDroneCornerCases}{None specified.}

\newcommand{\tempDroneSync}{No sample time stamps published. Asynchronous SUT and ReS both at \SI{25}{\hertz}; synchronization in post-processing through a mutually captured visual signal; mean resulting accuracy loss $\approx$\,\SI{0.14}{\meter}. \textit{Not published:} Once synchronized, SUT tracks were interpolated to the matched ReS time stamps. Thus, the closest overhanging SUT measurements were used, while overhanging ReS measurements were discarded.}
\newcommand{\tempDroneMatching}{Entire tracks are matched if their mean center distance (see above) during their non-occluded lifetimes $<$\,\SI{1.5}{\meter}.}
\newcommand{\tempDroneIncomplete}{FNs were not explicitly computed.}
\newcommand{\tempDroneDelays}{Temporally overhanging track parts are discarded by counting them as FNs/FPs. \textit{Not published}: Synchronization based on raw data time stamps (ReS) and time stamp of arrival at ROS driver node after firmware processing (SUT), respectively. ReS time stamps are subject to synchronization uncertainty (see above).}

\newcommand{\probDroneAOV}{SUT AOV has a fuzzy range, which is not explicitly modeled. Objects are considered irrespective of position within AOV.}
\newcommand{\probDroneObjDistance}{Deterministic distance function only}
\newcommand{\probDroneExistClass}{\textit{Not published:} Objects of any existence confidence were considered without any threshold. Only those parts of SUT tracks during which \textit{car} was the most likely classification were extracted and considered for matching.}

\begin{table*}[htbp]
	\fontsize{7.5pt}{7.5pt}\selectfont
	\begin{threeparttable}
		\centering
		\caption{Checklist applied to concrete test oracles for TPs/FPs/FNs}
		\label{table:case_study}
		
		\newcommand{\midrulewidth}{0.5pt}
		\newcommand{\asp}{1pt}
		\newcommand{\rotatedmultirow}[3]{%
			\parbox[t]{2mm}{\multirow{#1}{*}[#2]{\rotatebox[origin=c]{90}{#3}}}%
		}
		\begin{tabularx}{\linewidth}{
				>{\hsize=0.02\hsize}X 
				>{\hsize=0.35\hsize}X 
				>{\hsize=0.87\hsize}X 
				>{\hsize=0.87\hsize}X 
			}
			\toprule
			\multicolumn{2}{>{\hsize=\dimexpr0.4\hsize+0.4\tabcolsep+\arrayrulewidth\relax}X}{\textbf{Criterion}}                                    & \textbf{nuScenes tracking task \cite{nuScenesTracking2024}} & \textbf{Obtaining TPs from a drone\cite{Krajewski2020UsingDrones}}  \\ \midrule
			\rotatedmultirow{3}{-0.4cm}{\textit{Context}}                                & \scenariosCol                                             & \scenariosChallenge                                              & \scenariosDrone                                                     \\ \cline{2-4}
			\addlinespace[\asp]                                                          & \sutCol                                                   & \outChallenge                                                    & \outDrone                                                           \\ \cline{2-4}
			\addlinespace[\asp]                                                          & \furtherCol                                               & \furtherChallenge                                                & \furtherDrone                                                       \\ \midrule[\midrulewidth]
			\rotatedmultirow{8}{-2.5cm}{Basic (Sec. \ref{sec:oracle_simple})}            & \ref{sec:basic_aov_ref} Area of vision (AOV) of ReS            & \basicChallengeAOV                                               & \basicDroneAOV                                                      \\ \cline{2-4}
			\addlinespace[\asp]                                                          & \ref{sec:basic_occlusions} Perspective-related occlusions             & \basicChallengeOcclusion                                         & \basicDroneOcclusion                                                \\ \cline{2-4}
			\addlinespace[\asp]                                                          & \ref{sec:basic_ref_hw} ReS hardware                       & \basicChallengeReSHW                                             & \basicDroneReSHW                                                    \\ \cline{2-4}
			\addlinespace[\asp]                                                          & \ref{sec:basic_ref_labeling} ReS labeling                 & \basicChallengeReSLabling                                        & \basicDroneReSLabling                                               \\ \cline{2-4}
			\addlinespace[\asp]                                                          & \ref{sec:basic_areas} Relevant areas                      & \basicChallengeAreas                                             & \basicDroneAreas                                                    \\ \cline{2-4}
			\addlinespace[\asp]                                                          & \ref{sec:basic_spatial_alignment} Geometrical alignment      & \basicChallengeGeometrAlign                                      & \basicDroneGeometrAlign                                             \\ \cline{2-4}
			\addlinespace[\asp]                                                          & \ref{sec:basic_dist_func} Object distance function             & \basicChallengeObjDistance                                       & \basicDroneObjDistance                                              \\ \cline{2-4}
			\addlinespace[\asp]                                                          & \ref{sec:basic_multi_obj_matching} Multi-object matching    & \basicChallengeMultiObjMatching                                  & \basicDroneMultiObjMatching                                         \\ \cline{2-4}
						\addlinespace[\asp]                                                          & \ref{sec:basic_corner_cases} Corner cases    			& \basicChallengeCornerCases                               & \basicDroneCornerCases                                         \\ \midrule[\midrulewidth]
			\rotatedmultirow{4}{-1.2cm}{Temporal (Sec. \ref{sec:oracle_time})}           & \ref{sec:temp_sync} Temp\textit{}oral synchronization                       & \tempChallengeSync                                               & \tempDroneSync                                                      \\ \cline{2-4}
			\addlinespace[\asp]                                                          & \ref{sec:temp_matching} Matching in time                  & \tempChallengeMatching                                           & \tempDroneMatching                                                  \\ \cline{2-4}
			\addlinespace[\asp]                                                          & \ref{sec:temp_incomplete} Incomplete tracks               & \tempChallengeIncomplete                                         & \tempDroneIncomplete                                                \\ \cline{2-4}
			\addlinespace[\asp]                                                          & \ref{sec:temp_latency} Latency \& delays                  & \tempChallengeDelays                                             & \tempDroneDelays                                                    \\ \midrule[\midrulewidth]
			\rotatedmultirow{3}{-0.25cm}{Probabil. (Sec. \ref{sec:oracle_probabilistic})} & \ref{sec:prob_for} AOVs                                   & \probChallengeAOV                                                & \probDroneAOV                                                       \\ \cline{2-4}
			\addlinespace[\asp]                                                          & \ref{sec:prob_bbox} Object state \& distance function       & \probChallengeObjDistance                                        & \probDroneObjDistance                                               \\ \cline{2-4}
			\addlinespace[\asp]                                                          & \ref{sec:prob_thresholding} Object existence \& classification & \probChallengeExistClass                                         & \probDroneExistClass                                                \\ \bottomrule
		\end{tabularx}
	\end{threeparttable}
\end{table*}



\section{Conclusion}
\label{sec:conclusion}

The identification of TPs/FPs/FNs can be ambiguous due to uncertain reference data and unclear expectations of what a TP means. 
Since the literature had been lacking a comprehensive way to define such tests, their results have been difficult to compare.

This paper illustrates the functional aspects and implementation details of a test oracle for TPs/FPs/FNs.
It provides a condensed checklist that practitioners can follow to define and describe their own test oracles.

By following the present checklist, ambiguities in the oracle can be minimized. 
This makes TPs/FPs/FNs and the metrics based on them more comparable and better usable for task-oriented dependability claims.

The actual definition of a test oracle that precisely recalls perception events inducing vehicle-level failures remains part of ongoing research. 
Meanwhile, the presented checklist can assist in investigating the role of explicitly identified TPs/FPs/FNs in such oracles. 

\section*{Declarations}

\begin{acknowledgements}
	The author thanks Fabian Thomsen for providing the unpublished implementation details of our former publication. 
	Open Access funding enabled and organized by Projekt DEAL.
	The present work was conducted independently without related funding or employment. 
\end{acknowledgements}

\subsection*{Data availability}

The present paper is based only on publicly available literature and makes no quantitative argumentations based on data.
The underlying datasets of the two analyzed test oracles (Sec. \ref{sec:case_studies}) operate on the publicly available nuScenes dataset (\url{http://nuscenes.org}), and on a proprietary dataset from \cite{Krajewski2020UsingDrones}, respectively.

\subsection*{Conflicts of interest}

Michael Hoss declares no conflict of interest.


\bibliographystyle{spmpsci}      
\bibliography{AllSourcesMHO}

%
%

\end{document}